\RequirePackage{fix-cm}
\documentclass[twocolumn]{svjour3}          
\smartqed  
\usepackage{graphicx}
\usepackage{amsmath}
\usepackage{latexsym}
\usepackage{hyperref}
\usepackage{amsfonts}
\usepackage{color}
\usepackage{comment}
\usepackage[abs]{overpic}
\usepackage{tikz}   
\usepackage{subfigure}
\usepackage{array}
\usepackage{multicol}
\usepackage[normalem]{ulem}
\usepackage{bm}
\newcommand*{\bb}[1]{{{#1}}}

\usepackage{mathtools}
\DeclarePairedDelimiter\ceil{\lceil}{\rceil}
\usepackage[ruled,linesnumbered]{algorithm2e}
\newcommand{\SCH} {
Schr\"{o}dinger
}
\begin{document}
\title{\centering Sparse Approximation of 3D Meshes using the Spectral Geometry of the Hamiltonian Operator}
\titlerunning{Sparse Approximation of 3D Meshes using the Spectral Geometry of the Hamiltonian Operator}        
\author{\hspace{4cm} Yoni Choukroun \and Gautam Pai \and Ron Kimmel}
\institute{Technion-Israel Institute of Technology \at
              \email{(yonic,paigautam,ron)@cs.technion.ac.il}}

\maketitle
\begin{abstract}
The discrete Laplace operator is ubiquitous in spectral shape analysis, 
 since its eigenfunctions are provably optimal in representing smooth functions
 defined on the surface of the shape. 
Indeed, subspaces defined by its eigenfunctions have been utilized for shape compression, 
 treating the coordinates as smooth functions defined on the given surface.
However, surfaces of shapes in nature often contain geometric structures for which the 
 general smoothness assumption may fail to hold. 
At the other end, some explicit mesh compression algorithms utilize the order by which vertices 
 that represent the surface are traversed, a property which has been ignored in spectral approaches. 
Here, we incorporate the order of vertices into an operator that defines a novel spectral domain.
We propose a method for representing 3D meshes using the spectral geometry of the Hamiltonian operator, 
 integrated within a sparse approximation framework. 
We adapt the concept of a potential function from quantum physics and incorporate vertex ordering
 information into the potential, yielding a novel data-dependent operator. 
The potential function modifies the spectral geometry of the Laplacian to focus on regions with
 finer details of the given surface.
By sparsely encoding the geometry of the shape using the proposed data-dependent basis, we improve
 compression performance compared to previous results that use the standard Laplacian basis and 
  spectral graph wavelets.
\end{abstract}
\section{Introduction}
\begin{figure*}[tbp]
  \centering
  \mbox{}
  \includegraphics[width=\linewidth]{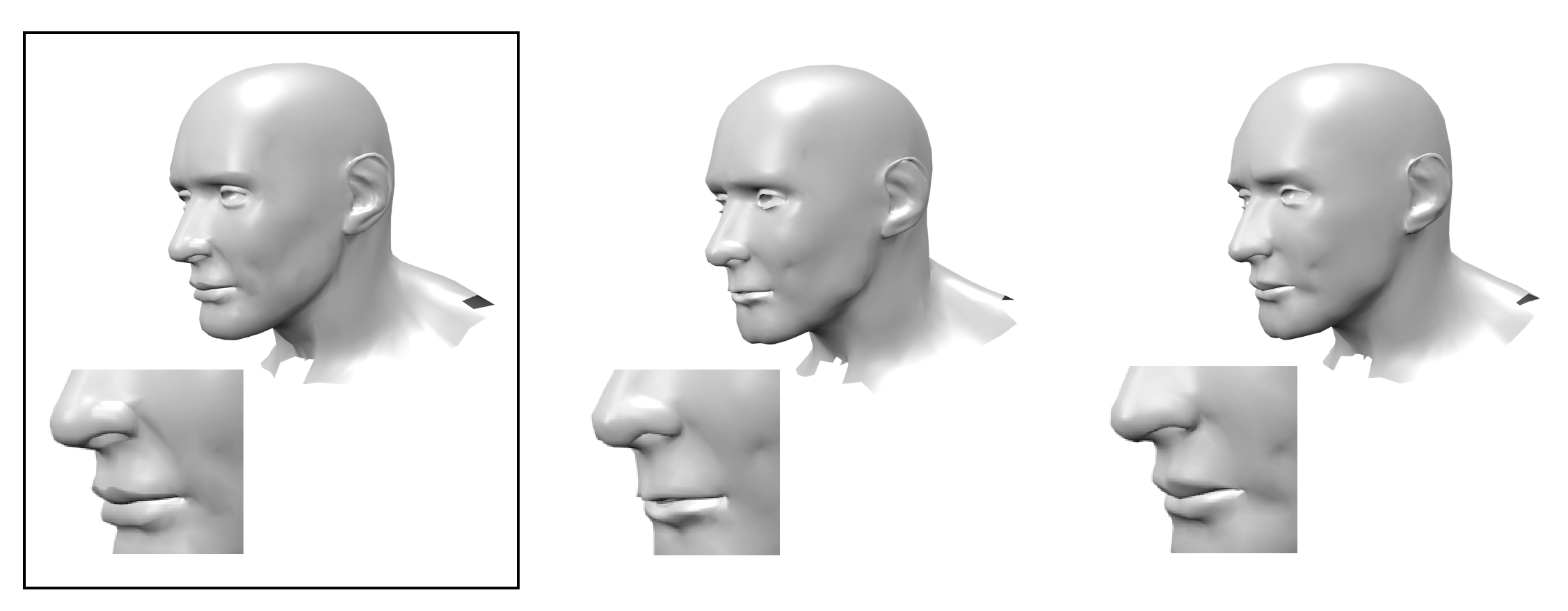}
  \hfill \mbox{}
  \caption{\label{fig:teaser}%
``The devil lies in the details.'' The boxed frame on the left is the original shape. The center and right figures show sparse approximations obtained using the eigenfunctions of the standard Laplacian and the Hamiltonian operators respectively, using a compression ratio of 1:10. The global Laplacian basis, also called manifold harmonics fail to capture the structure in the fine details. The approximation using the proposed Hamiltonian basis (figures on the right) preserves this significant fine structure for the same compression ratio. }
\end{figure*}
In most areas that involve representation of discrete virtual surfaces as 3D meshes, there has been an increasing trend in working with higher precision. 
This has led to the generation of meshes comprised of a large number of elements, for which the processing, visualization and storage of has become a challenge. 
The task of transmission of these geometric models over communication networks 
 can lead to a large amount of storage space and put a considerable strain on
 network resources. %

The research on mesh compression started in the direction of single-rate 
 compression methods that constructed a compact representation of an input
 mesh as a whole. 
However, the need for transmission of large geometric models led to the development of algorithms based on progressive mesh compression where a mesh can be transmitted and reconstructed progressively, that is, with increasing levels of detail. 
This enables the rendering of the 3D model in progression where the finer details 
 of the model can be processed in a cumulative fashion. 

The information contained in a mesh is generally divided into two categories: 
 the \textit{geometry} information, which is the position of each vertex of the mesh in the 3D Euclidean space and the \textit{connectivity} or \textit{topological} information, which describes the incidence relations between the mesh vertices.
Since the geometric information comprises a dominant part of the mesh, most recent algorithms focused on its efficient compression.
Complete reviews, summarizing work on mesh compression can be found in \cite{alliez2005recent,peng2005technologies,maglo20153d}.
\subsection{Related Results}
Fourier analysis has been used extensively in many signal processing areas.
By projecting the data into the frequency domain, one can retrieve a good approximation
 of the original signal with only a few Fourier coefficients.
A very prominent application is the JPEG \cite{wallace1992jpeg} image compression
 method based on the closely related 2D Discrete Cosine Transform. 

Karni and Gotsman \cite{karni2000spectral} proposed a generalization of the 
 Fourier basis on discrete graphs in order to compress mesh vertex positions. 
This is achieved by projecting the coordinate vectors onto the orthonormal basis obtained from the spectral decomposition of the combinatorial Laplacian of the shape.
For smooth meshes, most of the signal energy is contained in the low frequencies
 of the mesh spectrum and hence, the coefficients corresponding to the lower frequencies are sufficient to build a good approximation of the original mesh.
Ben-Chen and Gotsman \cite{ben2005optimality} proved that for certain geometric 
 models, spectral compression is optimal and is equivalent to Principal Component
 Analysis.

The idea of using a representative basis for approximating a mesh has been explored in detail in multiple prior works. 
 \cite{karni20013d} proposed to use a fixed basis independent of the mesh connectivity.
\cite{sorkine2005geometry} proposed a geometry-aware basis that considers both the connectivity and the geometry of the mesh for mesh compression.  
Mahadevan \cite{mahadevan2007adaptive} replaced the manifold harmonics by a 
 diffusion wavelet bases and showed an improvement in compression performance. 
This was attributed to the rich multi-scale nature of the wavelet basis.

Methods dealing with basis design for meshes have also been explored in other applications.
In order to overcome the non-local support of the Fourier eigenfunctions, compressed eigenfunctions 
 have been adapted from mathematical physics to shape analysis \cite{neumann2014compressed,bronstein2016consistent}.

Anisotropic phenomena, such as diffusion, has been explored in different  image processing applications \cite{weickert1998anisotropic} and has recently demonstrated impressive results in image compression \cite{peter2015optimised}. 
Anisotropic Laplacians have been introduced in order to improve the performance compared to the classical isotropic Laplace operator.
Kovnatsky et al. \cite{kovnatsky} modified the metric tensor of the manifold to take into account photometric information.
Andreux et al. \cite{andreux2014anisotropic} used principal curvatures in order to construct anisotropic tensor for the task of shape segmentation and shape matching. 

Keeping the anisotropic information out of the metric leads to lower order of derivatives, a property that yields numerically more robust operators.
Hildebrandt et al. \cite{hildebrandt2012modal} designed a new kind of eigenvibrations constructed as the Hessians of surface energies using extrinsic curvatures and deformation energies.
Similarly, Iglesias and Kimmel \cite{iglesias2012schrodinger} used artificial surface textures on shapes to define elliptic operators that give birth to a new family of diffusion distances. 
This operator, also called Hamiltonian, was recently explored in \cite{ChoukrounSK16}, which provides a broad theoretical foundation of the operator and its use in different shape analysis tasks, including data driven potential optimization, compressed manifold modes and shape matching. In this paper, we extend this work by analyzing the operator for the task of mesh compression and provide a powerful sparse representation framework based on the Hamiltonian operator, that outperforms previous spectral compression methods.


Given a basis, a plain spectral truncation of the signal in that basis is a fairly simple method for representation. Instead, sparse approximation techniques have been proposed and have become very popular.
The basic idea is to estimate a given signal as a \emph{sparse} linear combination of a large pool of constituent vectors - called a dictionary. These vectors, or atoms, are selected such that the coefficients of representation are sparse. 
The rationale is that high dimensional signals generally possess intrinsic structures that are better represented in a lower-dimensional linear subspace. 
Although these dictionaries have traditionally been populated with complete orthogonal bases, redundant or overcomplete dictionaries allow greater flexibility in design by better capturing the intrinsic characteristics of a signal.

The main difficulty with sparse algorithms is the availability of a rich representative dictionary. 
This seems trivial for signals defined over regularly and consistently sampled domains, like images and speech, but it is not straightforward to extend the idea to non-flat domains like meshes of surfaces in 3D or general graphs. 
The use of redundant representations for mesh representation and compression have started to emerge in
  \cite{peyre2005surface,tosic2006progressive,krivokuca2013sparse,zhong2014sparse}. 
We extend the ideas discussed in these papers by choosing the eigenfunctions of a data-aware operator as the constituents of a large overcomplete dictionary to represent the mesh.

\subsection{Contribution}
In this paper, we use the spectral geometry of the Hamiltonian operator for approximation of meshes representing surfaces in 3D. The operator is obtained by modifying the Laplacian with a potential function that defines the rate of oscillation of the harmonics on different regions of the surface.
We use a simple and efficient construction of the potential function using a vertex ordering scheme.
This modulates the Fourier basis of the 3D mesh to focus on crucial regions of the shape having fine geometric structures equivalent to high-frequencies in the Fourier domain. We optimize our approximation using a sparse representation framework by building optimized dictionaries in order to \emph{sparsely} encode mesh geometry. 

In contrast to \cite{ChoukrounSK16}, which suggests a simplistic truncation scheme for the representation of discrete surfaces, this paper provides an elaborate treatise on the the design of the Hamiltonian operator to tackle the specific task of spectral mesh compression.
We develop a novel multi-resolution dictionary learning mechanism which takes advantage of the adaptable nature of the Hamiltonian to design a context basis. 
The combination of the spectral geometry of the Hamiltonian in conjunction with a sparse approximation approach outperforms existing \emph{spectral} compression schemes as demonstrated in section \ref{section:results} and illustrated in Figure \ref{fig:teaser}.


\section{Background}\label{section:background}
\subsection{Laplace Beltrami Operator}
Consider a parametrized manifold
 $\mathcal{M} : \Omega  \subset \mathbb{R}^2 \to \mathbb{R}^3$ 
 with boundary $\partial\mathcal{M}$.
The space of square-integrable functions on $\mathcal{M}$ is denoted by 
 $L^{2}(\mathcal{M})=\{f:\mathcal{M}\rightarrow \mathbb{R} | \int_{\mathcal{M}}f^{2}da<\infty\}$ 
  with the standard inner product 
   $\langle f,g \rangle_{\mathcal{M}} = \int_\mathcal{M} fg \ da$, 
    where $da$ is an area element induced by the Riemannian metric
    $\langle \cdot, \cdot \rangle_{g}$.

The Laplace Beltrami Operator (LBO) acting on a scalar function 
  $f\in L^{2}(\mathcal{M})$, is defined as
\begin{equation}\label{eq:LBO}
\begin{aligned}
\centering
\Delta_{\mathcal{M}}f \equiv \mbox{div}_{\mathcal{M}}(\nabla_{\mathcal{M}} f),
\end{aligned}
\end{equation}
where $\nabla_{\mathcal{M}}$ represents the intrinsic gradient.
  The LBO is a self-adjoint operator and hence admits a spectral decomposition: 
  $\{\lambda_{i},\phi_{i}\}$, such that,
\begin{equation}\label{eq:LBOeigenspace}
\begin{aligned}
 -\Delta_\mathcal{M} \phi_i = \lambda_i \phi_i, \\
 \langle \phi_i ,\phi_j \rangle_\mathcal{M} = \delta_{ij},
 \end{aligned}
\end{equation}
where $\lambda_{i}\in\mathbb{R}$, $0=\lambda_{1}\le\lambda_{2}\le ...$  
and $\delta_{ij}$ the Kronecker delta function. In case $\mathcal{M}$ has a boundary, we add Neumann boundary condition to $\phi_{i}$. The LBO eigendecomposition can be obtained by minimizing the Dirichlet energy
\begin{equation}\label{eq:LBOFunctional}
\begin{aligned}
& \underset{\phi_{i}}{\text{min}}
& & \int_{\mathcal{M}} \|\nabla_{\mathcal{M}}\phi_{i} \|^{2}_{g} \ da, \ \\
& \text{s.t.}
& & \langle \phi_{i},\phi_{j}\rangle_{\mathcal{M}} = \delta_{ij}.
\end{aligned}
\end{equation}
Therefore, the LBO eigenfunctions can be seen as an
 extension of the \emph{smooth} Fourier harmonics in Euclidean space to manifolds and are also referred to as \textit{Manifold Harmonics}\cite{Levy06,ValletL08}.
\subsection{Graph Laplacian}
Let $\mathcal{G} = (\Pi, \Upsilon)$ be an undirected graph of $n$ vertices $\Pi = \{ \pi_{i}\}_{i=1}^{n}$ representing a sampled version of a manifold $\mathcal{M}$.
Let the graph edges be equipped with nonnegative weights $\{w_{ij}\}_{\{i,j\}=1}^{n}$, 
 defining the weighted adjacency matrix $\bb{W}$ of the graph as
\begin{equation}
\begin{aligned}
 \big[\bb{W}\big]_{ij} = \left 
  \{ \begin{array}{ll}
    \ \ \ w_{ij},  & \text{if}\  (i,j)\in \Upsilon,
     \cr
    \ \ \ 0,                                            & \text{ otherwise}.
  \end{array} \right .
\end{aligned}
\end{equation}
The diagonal vertex degree matrix $A$ is defined as
\begin{equation}
\begin{aligned}
 \big[\bb{A}\big]_{ii}=a_{ii}=\sum_{j}w_{ij}.
\end{aligned}
\end{equation}
The unnormalized graph Laplacian is defined as the 
 $n \times n$ matrix $\bb{L} = \bb{A}-\bb{W}$.
For properties of the Laplacian, the reader is referred to \cite{mohar1991laplacian,chung1997spectral}.

The combinatorial graph Laplacian is constructed by solely considering the connectivity of the mesh, 
 where the edge weights are defined as $w_{ij}=1$.
Therefore, the degree matrix $\bb{A}$ is populated with the valence of each vertex.
While other \emph{geometric} discretizations of the Laplacian on graphs exist
 \cite{pinkall1993computing}, the combinatorial Laplacian is used in this paper, 
  since the mesh topology can be efficiently encoded . 

The discrete manifold harmonics are then obtained by solving the spectral decomposition 
 of the combinatorial Laplacian 
\begin{equation}
 \bb{L}\bb{\Phi}=\bb{\Phi}\Lambda.
\end{equation}
Here, $\bb{\Phi}\in \mathbb{R}^{n\times n}$ is the matrix of eigenvectors where each
 column $\bb{\phi}_{k}$ is one eigenvector and $\Lambda$ is the diagonal matrix of eigenvalues where each element $\lambda_{k}$ is one eigenvalue.

\subsection{Spectral Mesh Compression}
\begin{figure}[htbp]
\begin{overpic}[width=1\linewidth,trim={180 250 130 35},clip]{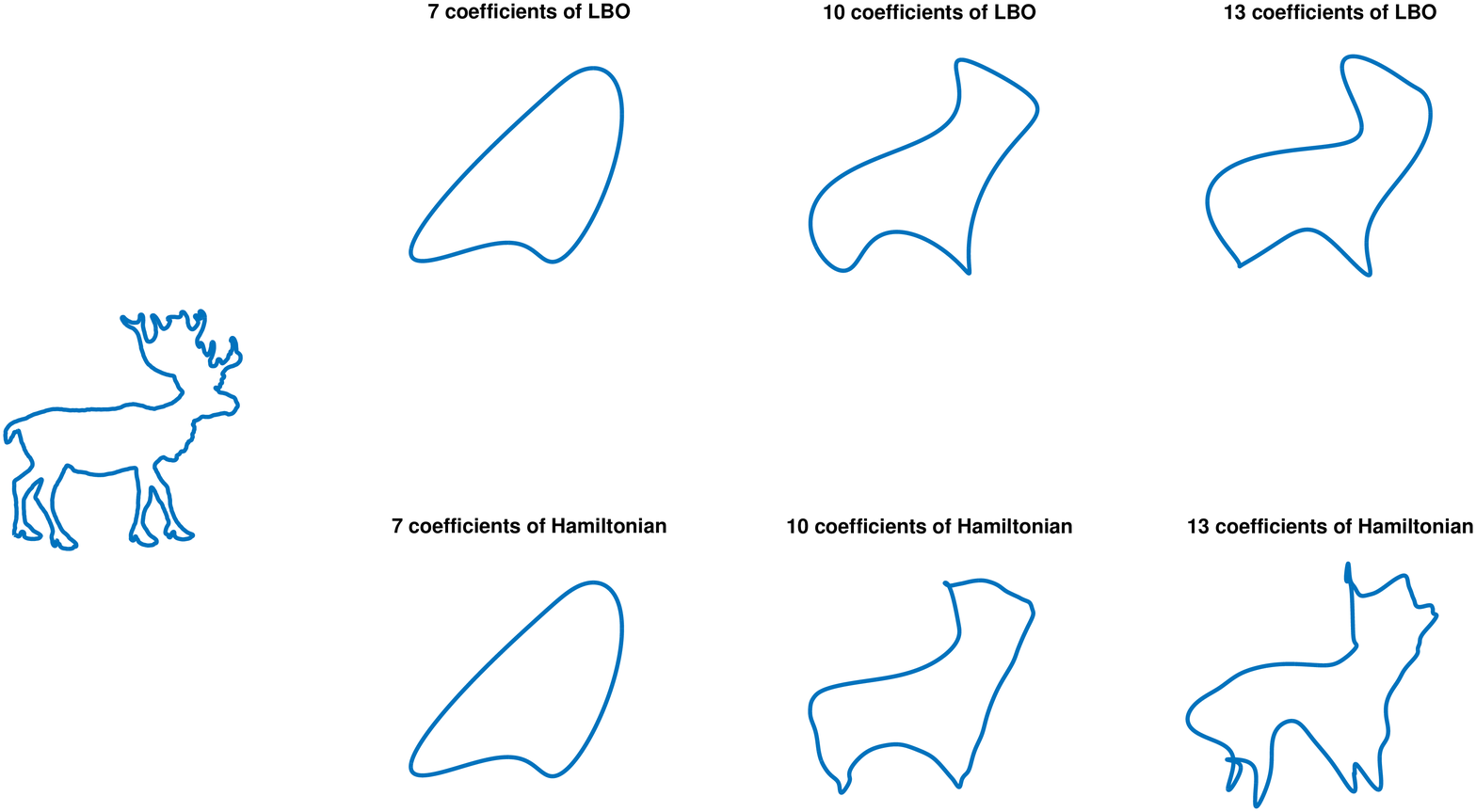}
\centering
 \end{overpic}
 \caption{\label{fig:truncation_curves} 
Approximation of a planar curve by simple truncation of the LBO and the proposed basis (Hamiltonian). 
 The potential function for the Hamiltonian is defined by the error between the curve and its 
  LBO truncated approximation. 
 By biasing the 1D harmonics with this potential, the reconstruction with the Hamiltonian can be
  seen to preserve finer details of the contour for the same number of coefficients.}
 \end{figure}

Given a complete basis $\{\varphi_{i}\}_{i=0}^{\infty}$ on $\mathcal{M}$, any function 
 $f \in L^{2}(\mathcal{M})$ can be expressed as
\begin{equation}\label{eq:representation}
\begin{aligned}
f = \sum_{i=1}^{\infty}\langle f,\varphi_{i} \rangle_{\mathcal{M}} \varphi_{i}=
 \sum_{i=1}^{\infty}\hat{f}_{i} \varphi_{i},
\end{aligned}
\end{equation}
 where $\hat{f}_{i}$ is the i-th (manifold) Fourier coefficient of $f$.
Since the manifold harmonics form a complete basis, Karni and Gotsman \cite{karni2000spectral} proposed to use them for spectral mesh compression. 
Considering the mesh coordinate functions $\bb{X}, \bb{Y},\bb{Z}$
 as functions 
 defined on the vertices of the mesh, the basic idea of spectral mesh compression is to compute 
  the Fourier transform of the coordinate functions and truncate the high-frequency coefficients.
Thus, if $\bb{u}$ is a coordinate function, it can be approximated using the graph Laplacian 
 decomposition as
\begin{equation}\label{eq:representation_k}
\begin{aligned}
\bb{u} \approx \sum_{i=1}^{K}\langle \bb{u},\phi_{i} \rangle_{\mathcal{G}} \phi_{i}
\end{aligned}
\end{equation}
with $K\ll n$ and
 $\langle \bb{u},\phi_{i} \rangle_{\mathcal{G}}=\phi_{i}^{T}\bb{u}$.
By using the graph connectivity information and constructing the manifold harmonics as 
 a representative basis, the geometric information of the mesh can be encoded efficiently 
  through the obtained coefficients. 
Figure \ref{fig:truncation_curves} illustrates spectral reconstructions as applied to a toy example of a 2D curve.

Simple truncation as showed in eq. (\ref{eq:representation_k}) uses a restrictive assumption of only focusing on the lower frequencies and is analogous to low-pass filtering of the signal.  
In general, the lower frequencies encapsulate most of signal energy and the reconstructed mesh tends to preserve the global features of the original. 
However, local geometry and fine details of the mesh corresponding to high frequencies are generally missing and require a high number of eigenvectors for their preservation if a truncation approach is followed. 

In the next section we propose a more pragmatic approach of sparsity where the data itself searches for its best constituents (frequencies) with the objective of attaining sparsity in a certain basis of representation. This leads to an improved compression strategy. 

\subsection{Sparse Approximation of Mesh Coordinates}
As motivated in the last section, classical methods are based on the premise that family of signals are 
 best represented using only the first few components of an optimal orthonormal basis.
The concept of sparsity provides an alternative perspective for representation.

Consider a graph $\mathcal{G}$ and $\bb{D}\in \mathbb{R}^{n\times m}$ a normalized overcomplete
 ($m\gg n$) dictionary containing $m$ atoms $d_{i}\in \mathbb{R}^{n}$. 
We aim to approximate any given signal $\bb{u}$ on $\mathcal{G}$ by
\begin{equation}\label{eq:representation_dic}
\begin{aligned}
\bb{u} = \bb{D}\alpha = \sum_{i=1}^{m}d_{i}\alpha_{i},
\end{aligned}
\end{equation}
 with $\alpha = \big[\alpha_{1},...,\alpha_{m}\big]^{T}$ the coefficient representation of the
 input signal $\bb{u}$ with respect to the dictionary $\bb{D}$.
In order to achieve significant reduction in storage we should assume that the number of
 non-zero elements of the coefficient vector $\alpha$ should satisfy $\|\alpha\|_{0}=k\ll n$. 
Thus, the sparse approximation of the signal can be obtained by solving the sparse coding problem
\begin{equation}
\begin{aligned}
  & \min_{\alpha} \|\bb{u} - \bb{D}\alpha\|^{2}_{2}\\
  & \text{s.t.} \ \ \|\alpha\|_{0}=k.
\end{aligned}
\end{equation}

Let us be given a multiple channels signals $\bb{U} \in \mathbb{R}^{n\times c}$,
 the new coefficient matrix is designated by $\Gamma=\big[\alpha_{1},...,\alpha_{c}\big]^{T}$
  with $\bb{U}=\bb{D}\Gamma$. 
The new sparse approximation is obtained by solving the simultaneous sparse approximation
\begin{equation}\label{eq:simultaenous_sparse}
\begin{aligned}
  & \min_{\Gamma} \|\bb{U} - \bb{D}\Gamma\|^{2}_{F} \\
  & \text{s.t.} \ \ \|\alpha_{1}\|_{0}=...=\|\alpha_{c}\|_{0}=k,
\end{aligned}
\end{equation}
 where $\|\cdot \|_{F}$ denotes the Frobenius norm.
In our mesh compression configuration, the input signal is defined by the mesh coordinates
 such that $\bb{U}=\big[\bb{X},\bb{Y},\bb{Z}\big]$.

Since sparse coding solution is a combinatorial NP-hard problem, well known relaxations are generally 
 used \cite{mallat1993matching,pati1993orthogonal}. 
In this paper Simultaneous Orthogonal Matching Pursuit algorithm \cite{tropp2006algorithms} is used for sparse approximation of the mesh geometry.
Pursuit algorithms are based on a greedy approach of solving for the optimization in (\ref{eq:simultaenous_sparse}) by incrementally selecting atoms from the dictionary $\bb{D}$ that minimize the residual error. The orthogonal matching pursuit is one such way of picking the atoms where at each stage of the selection, the residual error vector is orthogonal to the atoms already selected in the support and hence will not be chosen again.
Simultaneous OMP is a scheme of atom selection such that multiple signals \emph{have the same support} in the dictionary $\bb{D}$, i.e. simultaneously solving three sparse problems by imposing same support constraint.

The success of a sparsity based algorithm depends on the availability of a representative dictionary.
However, developing dictionaries for the representation of functions defined over graphs is a 
 non-trivial task, since they do not possess a fixed structure as in Euclidean domains (grid).
As a result, elements composed from fixed spectral bases have been used for populating the atoms 
 of the dictionary \cite{krivokuca2013sparse}. 

The manifold harmonics can be used because of their stability and compactness, that is, 
 low frequencies contain most of the signal energy.
However, fine local structures are not well represented using this basis due to the global
 nature of their support.
In \cite{zhong2014sparse}, it was shown that a dictionary constructed from spectral graph 
 wavelets \cite{hammond2011wavelets} show better compression performance as compared to the 
  standard Laplacian basis, since it encapsulates fine details and some dominant high-frequency
  structures of the mesh.
However, since the locality of the basis cannot be controlled, the wavelet dictionaries are 
 highly overcomplete, yielding large dictionaries which enforce higher compression rates.
These shortcomings motivate the need for finding a basis which can be modulated in a data-dependent manner. This will lead to an optimal
local-global structure trade-off, thereby requiring a smaller cardinality of representation. 
In the next section we borrow an operator from quantum mechanics which has precisely the aforementioned properties. 

\section{Hamiltonian Operator}\label{section:hamiltonian}
\subsection{Definition}
\begin{figure}[htbp]
\begin{overpic}[width=1\linewidth]{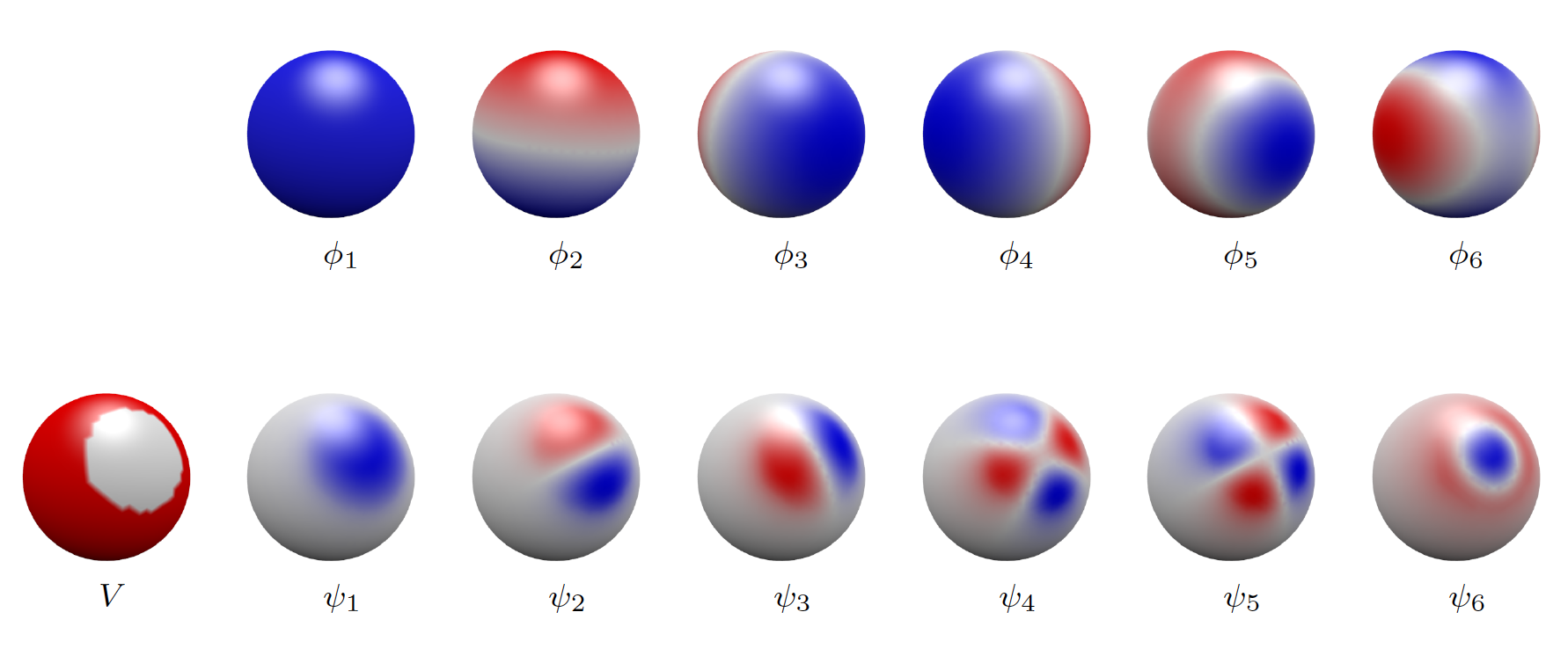}
\centering
 \end{overpic}
 \caption{\label{fig:spheres} 
 First eigenfunctions of the LBO on a sphere (top). Potential function defined on the sphere and the corresponding Hamiltonian basis (bottom). Hot and cold colors represent positive and negative values respectively. The Hamiltonian basis concentrates the harmonics to the low potential region.}
 \end{figure}

A Hamiltonian operator $H$, also called \SCH operator, is an operator acting on a scalar 
 function $f\in L^{2}(\mathcal{M})$ on a manifold $\mathcal{M}$ that has the form 
\begin{equation}\label{eq:Hamiltonian}
Hf=-\Delta_{\mathcal{M}}f + \mu Vf,
\end{equation}
 where $V:\mathcal{M}\rightarrow \mathbb{R}$ is a potential function and $\mu \in \mathbb{R}$. 
The Hamiltonian plays a fundamental role in the field of quantum mechanics via the 
 famous Schr\"{o}dinger equation that describes the wave motion of a particle.
The spectral decomposition $\{\psi_{i},E_{i}\}_{i=0}^{\infty}$ of the Hamiltonian is defined by 
\begin{equation}\label{eq:Hamiltonian_eigen}
\begin{aligned}
H\psi_{i} = E_i \psi_{i}
\end{aligned}
\end{equation}
 where $E_{i}$ denotes the energy of a particle at the stationary eigenstate $\psi_i$.
Here, $\psi_{i}(x)$ 
  represents the wave function of the particle such that $|\psi_{i}(x)|^2$ is interpreted as the 
  probability distribution of finding the particle at a given position $x$.
Since the Hamiltonian is a symmetric operator, its eigenfunctions form a complete orthonormal basis
 on the manifold $\mathcal{M}$. 
An illustration of the Hamiltonian basis is given in figure \ref{fig:spheres}.

The Hamiltonian eigendecomposition from (\ref{eq:Hamiltonian_eigen}) is obtained from the 
 Euler-Lagrange equation of
\begin{equation}\label{eq:SchFunctional}
\begin{aligned}
& \underset{\psi_{i}}{\text{min}}
& & \int_{\mathcal{M}} 
 \left (\|\nabla_{\mathcal{M}}\psi_{i} \|^{2}_{g} +\mu V \psi_{i}^2\right )  da, \cr
& \text{s.t.}
& & \langle \psi_{i},\psi_{j}\rangle_{\mathcal{M}} = \delta_{ij}.
\end{aligned}
\end{equation}

The parameter $\mu$ controls the potential energy and defines the trade-off between local 
 and global support of the basis. 
Larger values of $\mu$ will give solutions that concentrate on the low potential regions, 
 while a smaller $\mu$ will give solutions that better minimize the total energy at the expense of more extended wave functions.

In its discrete setting, the Hamiltonian basis is obtained by solving 
\begin{equation}
\bb{H}\bb{\psi_{i}}=(\bb{L}+\mu \bb{V})\bb{\psi_{i}}=\bb{E_{i}}\bb{\psi_{i}}
\end{equation}
 with $\bb{\{\psi_{i}}\}_{i=1}^{n}\in \mathbb{R}^{n}$. 
Here $\bb{L}$ denotes the previously described graph Laplacian, and $\bb{V}$ a diagonal matrix that 
 is defined by the potential scalar function values at vertices $\pi_{i} \in \Pi$.
Since $\bb{V}$ only modifies the diagonal of the sparse Laplacian, the Hamiltonian is also a 
 sparse matrix with the same non-zero entries. 
Thus, there is no increase in the computational cost of the decomposition compared to that of the Laplacian.

\subsection{Analysis and Visualization of the Operator}
Among the numerous reasons that motivated the selection of the Laplacian basis for shape analysis, one ubiquitous
is its efficiency in representing smooth functions. 
This property is defined by the following observation
for functions with $L^2$ bounded gradient magnitudes:

\begin{equation}\label{eq:represent}
\begin{aligned}
&\|r_n\|_{2}^{2}=\left\|f-\sum_{i=1}^{n}\langle f,{\phi}_{i}\rangle {\phi}_{i}\right \|_{2}^{2}\leq \frac{\|\nabla_g f \|^{2}_{g}}{\lambda_{n+1}}.
\end{aligned}
\end{equation}
This result can be established as follows,
\begin{equation}\label{eq:represent}
\begin{aligned}
&\|\nabla_{g}f\|_{g}^{2}=\int_{\mathcal{M}}(-\Delta_{\mathcal{M}} f)f \  da \\ &=\sum_{i=1}^{\infty}\int_{\mathcal{M}}(\langle f,\phi_{i}\rangle_{\mathcal{M}}\lambda_{i}\phi_{i}) f \ da = \sum_{i=1}^{\infty}\lambda_{i}\langle f,\phi_{i}\rangle_{\mathcal{M}}^{2} \\ &\geq 
\sum_{i=n+1}^{\infty}\lambda_{i}\langle f,\phi_{i}\rangle_{\mathcal{M}}^{2} \geq 
\lambda_{n+1}\sum_{i=n+1}^{\infty}\langle f,\phi_{i}\rangle_{\mathcal{M}}^{2}=\lambda_{n+1}\|r_{n}\|^{2}_{2}.
\end{aligned}
\end{equation}
The representation error $r_n$ for a function $f$ using the Laplacian basis $\{ \bb{\phi}_{i} \}_{i=1}^{i=\infty}$ is bounded by a factor of the squared magnitude of its gradient. This result was subsequently proved to be optimal for representing smooth functions over surfaces in \cite{aflalo2015optimality}, which says that there exists no other basis with such a bound for all possible $L^2(\mathcal{M})$ bounded gradient functions.

\begin{figure}[htbp]
\begin{overpic}[width=1\linewidth,trim={180 230 120 200},clip]{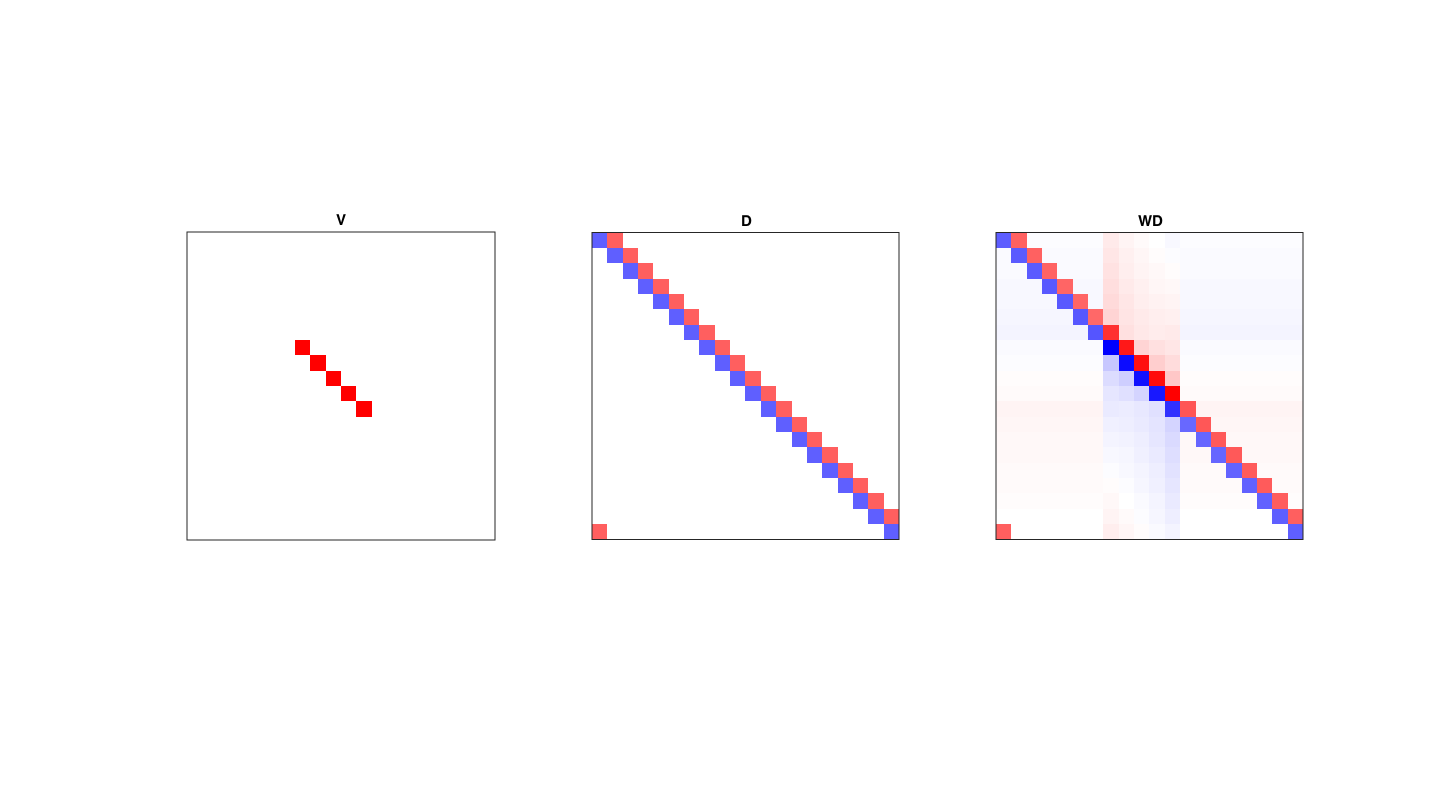}
\centering
 \end{overpic}
 \caption{\label{fig:OptimalityMat} Matrix visualizations of the potential function (left), the standard gradient matrix D (middle) and the weighted gradient matrix $WD = (I+D^{-T}VD^{-1})^{\frac{1}{2}}D$ (right) for Euclidean 1D example. 
}
\end{figure}

To gain further insight, we simulate this result in a discrete setting by analyzing the gradient operators for a simple one dimension case. 
In matrix form, since $L$ is a positive semi-definite matrix, there exists a matrix $D$ such that $\bb{L} = D^{T}D$ and the discrete equivalent of result \ref{eq:represent} is: 
\begin{equation}\label{eq:represent_discrete}
\begin{aligned}
\|r_n\|_{2}^{2}=\left\|f-\sum_{i=1}^{n}\langle f,\bb{\phi}_{i}\rangle \bb{\phi}_{i}\right \|_{2}^{2}\leq \frac{\|Df \|_{2}^{2}}{\lambda_{n+1}},
\end{aligned}
\end{equation}
It is interesting to perform a similar analysis on the eigenfunctions of the Hamiltonian and Figure \ref{fig:OptimalityMat} shows a visualization of the gradient operator of the Laplacian and the induced gradient of the Hamiltonian.
The result in equation (\ref{eq:represent_discrete}) gets modified to account for the action of the potential.
\begin{equation}\label{eq:represent2}
\begin{aligned}
\|r_n\|_{2}^{2}=\left\|f-\sum_{i=1}^{n}\langle f,\bb{\psi}_{i}\rangle \bb{\psi}_{i}\right \|_{2}^{2}\leq \frac{\|\bb{W}Df \|_{2}^{2}}{\bb{E}_{n+1}},
\end{aligned}
\end{equation}
with $\bb{W} = (\bb{I}+{D}^{-T}\bb{V}D^{-1})^{\frac{1}{2}}$ such that $\bb{H} = (\bb{W}D)^{T}\bb{W}D$. Thus the Hamiltonian basis can be considered as optimal for the representation of functions which have a bounded \emph{weighted} gradient. 
The weights depend on the potential function and its action in different regions.

The gradient matrix $D$ for the 1D case is the result of a simple finite difference scheme of one dimensional forward differentiation yielding a circulant populated with the kernel $[1, -1]$ circulated along the diagonal. 
The Hamiltonian has the effect of escalating the cost of a strong gradient \emph{exclusively in regions of high potential}. 
This shows that coefficients of the induced gradient matrix are larger in the high potential regions as compared to anywhere else in the domain.
The non-zero values of the off-diagonal elements suggest that the linear operation loses its shift invariance property and the effect of a derivative is no more an exclusive property of its local neighborhood but also depends on its global positioning imposed by the potential function.  
Thus the Hamiltonian operator advocates measuring smoothness differently for different regions of the domain and this is a useful property to exploit in a compression setup. 

The Hamiltonian operator can improve the compression performance by modifying the harmonics in order to emphasize \textit{designated} regions of interest.
We propose a method for choosing a potential function which does not require any additional encoding.  
We construct a dictionary based on the eigendecomposition of both the Laplacian and the Hamiltonian 
 that can focus on difficult reconstruction areas of the harmonics. 
Thus, by designing high vibrations in selected regions of the shape, our dictionary is much 
 less redundant than the wavelets proposed in \cite{mahadevan2007adaptive}, has a better 
 ability to encode high frequency details and achieves better compression performance.

\section{Hamiltonian Operator for Spectral Mesh Compression}\label{section:algorithm}
\subsection{Potential Design}
\label{potenial_design}
Consider a mesh (graph) $M=\{\Pi,\Upsilon \}$ with $|\Pi|=n$ and the coordinates matrix $\bb{U}$. 
Given the Laplacian eigenvectors matrix $\Phi \in \mathbb{R}^{n \times n}$, one can design a fixed 
 dictionary $\bb{D}_{L}$ composed of the full harmonics basis matrix $\bb{D}_{L}=\Phi$ and solve the 
  simultaneous sparse coding problem of (\ref{eq:simultaenous_sparse}) 
   where $\tilde{\bb{U}} = \bb{D}_{L}\Gamma \approx \bb{U}$.
Then, a $L_{2}$ norm reconstruction error $\epsilon(\pi_{i})=\|\pi_{i}-\hat{\pi}_{i}\|_{2}$ 
 can be obtained at each vertex of the mesh.
This vertex error profile over the mesh provides an indication of difficult regions that 
 can be enhanced through the potential.

Since a fixed potential must be used to avoid additional encoding, we permute the arbitrary 
 indexing of the mesh according to this error. 
The vertices are sorted in ascending order such that vertices having small indexes correspond 
 to points with a small error while vertices with the larger index correspond to regions with 
 higher distortion in approximation. 
It can be done trivially in $\mathcal{O}(\Upsilon)$ complexity and has no influence on the mesh
 itself since the ordering has no physical meaning. 

\begin{figure}[htbp]
\begin{overpic}[width=1\linewidth,trim={160 80 130 35},clip]{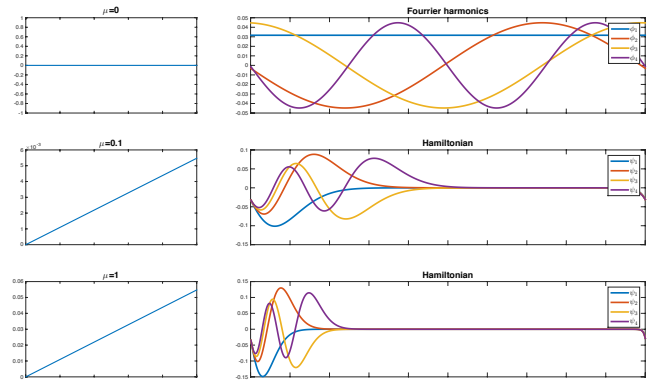}
\centering
 \end{overpic}
 \caption{\label{fig:linearPotential} Zero potential (left up) and its corresponding harmonics (right up). Linear potential for different parameter $\mu$ (left middle and bottom) and their corresponding 
  first eigenfunctions (right middle and bottom).
  The Hamiltonian eigenfunctions focus on the low potential areas without neglecting other regions for higher energies.
}
 \end{figure}

Given the Hamiltonian $\bb{H}=\bb{L}+\mu \bb{V}$, we propose to use a fixed potential matrix defined as
\begin{equation}\label{eq:potentialDef}
\begin{aligned}
\bb{V} = \text{diag}(1,..,n),
\end{aligned}
\end{equation}
 normalized such that $\|\bb{V}\|_{F}^{2} = \sum _{i}^{n}\bb{V}_{ii}^2 = 1$.
We illustrate the influence of a linear potential in one dimension in Figure \ref{fig:linearPotential}. 
Since we sorted the indexes according to the error, a designed hierarchical energy of the potential 
 is defined over the shape while the potential encoding itself is avoided.
At the decoder side, the Hamiltonian can be built using the mesh connectivity (Laplacian) and
 the encoded scalar $\mu$ coupled with the fixed linear potential $\bb{V}$ described above. 

\subsection{Dictionary Construction}
We construct a dictionary built from the eigendecomposition of both the Laplacian and the Hamiltonian
 to benefit from their global and local properties. 
The potential is defined according to the representation error of the Laplacian basis as described 
 in the previous section.
We encode multiple constants $\mu$ in order to obtain a multi-resolution of the basis.
The regularization constants are found via a direct search on a given domain.

Thus, our dictionary can be obtained as:
\begin{equation}
\bb{D}_{S}=[\Phi,\Psi_{\mu_{1}},...,\Psi_{\mu_{S}}],
\end{equation}
where the sub-dictionary $\Psi_{\mu_{i}}\in \mathbb{R}^{n \times n}$ is the matrix of eigenvectors
 of the Hamiltonian $\bb{L}+\mu_{i}\bb{V}$ and $\Phi=\Psi_{0}$.
The constant $\mu_{i}$ is obtained by solving
\begin{equation}\label{eq:mu_solution}
\begin{aligned}
  \min_{\mu_{i}} \ \ & \min_{\Gamma} \|\bb{U} - \bb{D_{i}}\Gamma\|^{2}_{F} \\
  & \text{s.t.} \ \ \|\alpha_{1}\|_{0}=...=\|\alpha_{c}\|_{0}=k,
\end{aligned}
\end{equation}
The algorithm is summarized in Algorithm \ref{alg:main_alg}.
Given a compression ratio requirement, the number of atoms is calculated according to the formulas elucidated in Section \ref{sec:comp_ratio}.

\begin{algorithm}[]\label{alg:main_alg}
\SetAlgoLined
\SetKwInOut{Input}{Input}\SetKwInOut{Output}{Output}
\SetKwRepeat{Do}{do}{while}%
\Input{Mesh connectivity $\Upsilon$, mesh coordinates $\bb{U}$ and compression ratio $k$}
\Output{Sparse coding $\{\alpha_{i}\}_{i}$ and coefficients $\{\mu_{j}\}_{j}$}
\BlankLine
Compute the pointwise error $\epsilon(\pi_{i})=\|\tilde{\bb{U}}(\pi_{i})-\bb{U}(\pi_{i})\|_{2}$ 
 where $\tilde{\bb{U}} = \Phi\Gamma$ is obtained from (\ref{eq:simultaenous_sparse});\\
 Permute the vertices according to the sorting of the error $\epsilon$;\\
 Build constant potential $\bb{V}$ as in (\ref{eq:potentialDef}) and set $j=0$;\\

  \Do{Decrease in representation error}{
  j=j+1;\\
    Find $\mu_{j}$ minimizing  (\ref{eq:mu_solution}) using $\bb{D}_{j}=[\Phi,\Psi_{\mu_{1}}...,\Psi_{\mu_{j}}]$;
    }
 \caption{Algorithm for sparse approximation}
\end{algorithm} 
\subsection{Discussion}
Algorithm \ref{alg:main_alg} is designed to take advantage of the two principle components of our contribution: sparsity and data-dependent basis. The potential design described in section \ref{potenial_design} alters the Laplacian basis to focus on error-prone regions of the shape. The extent to which such an alteration must be enforced is encoded in the coefficients: $\{\mu_{j}\}_{j=1}^{j=K}$ which are sequentially optimized in Algorithm \ref{alg:main_alg}. The final dictionary is comprised of the regular Laplacian eigenvectors appended with these Hamiltonian eigenvectors. The results in section \ref{section:results} demonstrate that this provides a distinct improvement in shape-reconstruction errors.

Another advantage of our approach is that the potential design is based on mesh connectivity information alone. 
The vertex re-ordering enables us to encode information about the error-prone regions of the shape, which are encoded in the eigenvectors of the Hamiltonian.
This re-ordering scheme avoids any secondary encoding of information and hence provides a distinct gain in saving of information units for compression.
 
\section{Experimental Results}\label{section:results}
\subsection{Evaluation Metric}
In order to measure the loss resulting from an estimated approximation, 
 Karni and Gotsman \cite{karni2000spectral} used a visual metric that captures the visual
  difference between the original mesh $M=\{\Pi,\Upsilon\}$ 
  and its reconstruction $\tilde{M}=\{\tilde{\Pi},\tilde{\Upsilon}\}$.
The measure is a linear combination of the RMS geometric distance between corresponding vertices
 in both models and a visual metric capturing the smoothness of the reconstruction.
The error $\epsilon \in \mathbb{R}^{n}$ at each vertex is defined as 
\begin{equation}\label{eq:karni_metric}
\begin{aligned}
\epsilon_{M} (\tilde{\pi}_{i}) = \frac{1}{2n}\bigg(\|\pi_{i} -\tilde{\pi}_{i}\|_{2}+\|GL(\pi_{i}) -GL(\tilde{\pi}_{i})\|_{2} \bigg).
\end{aligned}
\end{equation}
The visual metric $GL$ is defined as
\begin{equation}
GL({ \pi_{i}}) = { \pi_{i}} - \frac{\sum_{j \in \mathcal{N}(i)} l_{ij}^{-1} \; { \pi_{j} }}{\sum_{j \in \mathcal{N}(i)}\; l_{ij}^{-1}}
\end{equation}
 with $\mathcal{N}(i)$ is the set of indexes of the neighbors of vertex $i$, and $l_{ij}$
 is the geometric distance between vertices $i$ and $j$. 
A global representation error is obtained by summing $\epsilon_{M}$ over the mesh vertices.
We normalize the error by the area of the shape for uniform comparison between different shapes.
\subsection{Compression Ratio}\label{sec:comp_ratio}
\begin{figure*}[tbp]
  \centering
  \mbox{}
  \includegraphics[width=\linewidth]{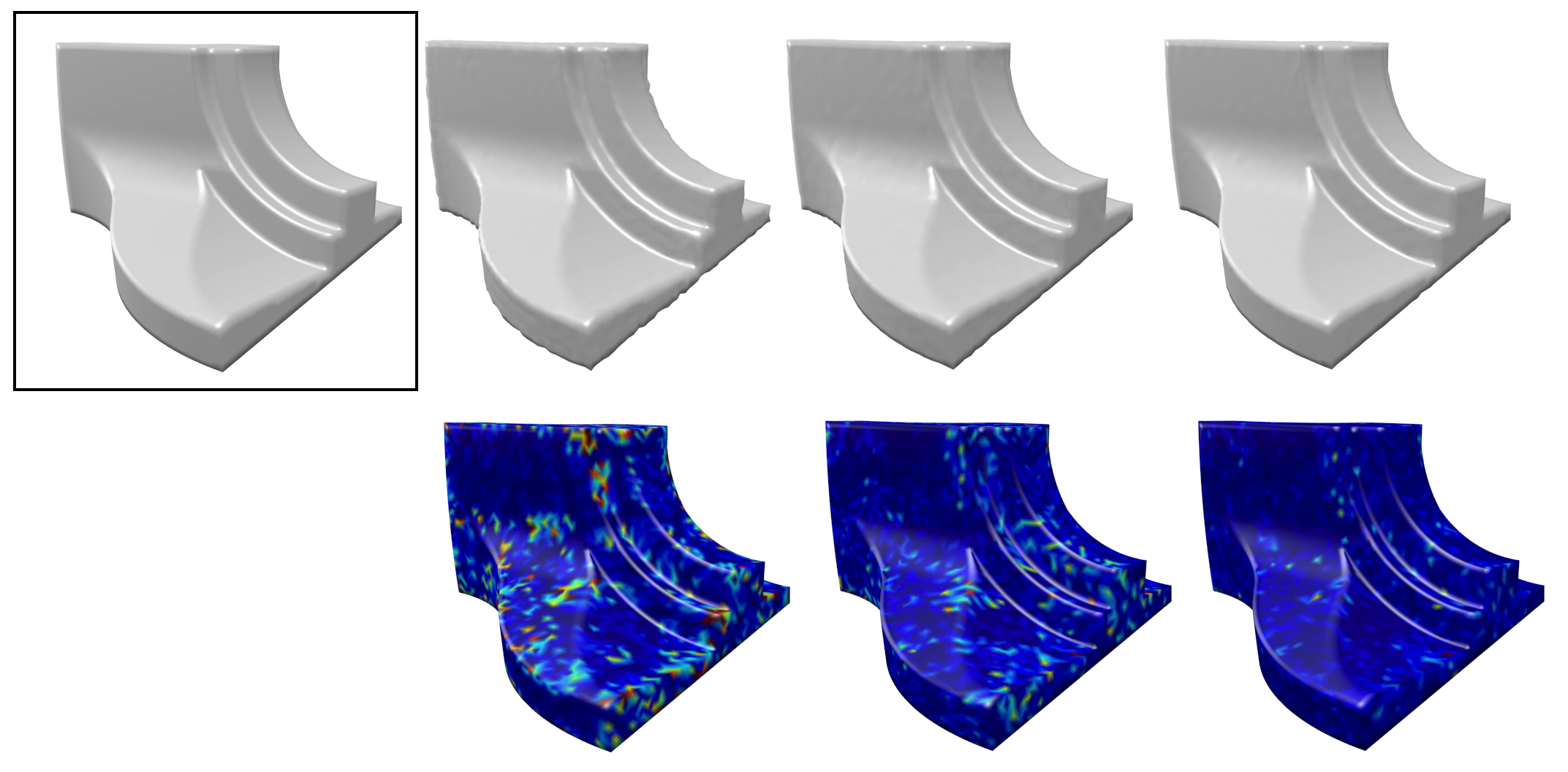}
  \hfill \mbox{}
  \caption{\label{fig:fandiskRecon}%
Fandisk model (boxed mesh) composed of 12946 vertices.(Top) Shape reconstruction for a compression ratio of $6:10$ using dictionaries composed of the MHB, MHB+SGW and MHB+Hamiltonian (from left to right).
(Bottom) Reconstruction error using the proposed spectral bases.
Cold and hot colors represent non-negative low and high values respectively.
 }
\end{figure*}

The compression ratio calculation is obtained as a ratio of the net information before and 
 after the dictionary encoding \cite{zhong2014sparse}.
If each coordinate takes up to $k$ bits, then for a mesh of $n$ vertices, the total uncompressed 
 information is $3nk$. 
Now, assuming that the dictionary $\bb{D} \in \mathbb{R}^{m\times n}$ is known in advance, or can be constructed from the connectivity information of the mesh, on both the
 encoder and decoder sides, let $n_d$ be the total number of selected atoms each taking $k_d$ bits to encode 
  each coordinate function, making the total information equal to $3n_d k_d$.  
The information required to store the indexes of the coefficients which is given by 
 $\min(m,n_d \ceil{\log_2m})$, The compression ratio  is given as
\begin{equation}\label{eq:CR_eq}
 C.R. = \frac{3n_d k_d+\min(m,n_d \ceil{\log_2m})}{3nk}.
\end{equation}
The $\min$ function is used since it can be more efficient to represent the selected atoms with a bit vector of size $m$ instead of the indexes themselves.
The compression ratio in the coefficient truncation scheme is given by
\begin{equation}
C.R. = \frac{n_d}{n},
\end{equation}
 since only the ordered coefficients must be encoded.
The compression ratio of the Hamiltonian dictionary can be easily extended by adding $n_{\mu}k_{\mu}$ to the numerator of  (\ref{eq:CR_eq}), with $n_{\mu}$ the number of regularization coefficients of the Hamiltonian and $k_{\mu}$ the number of bits required to encode each one of them.


\subsection{Shape partition}
Spectral methods, including the one proposed in this paper are hinged on the eigendecomposition of a large matrix.
This $\mathcal{O}(n^{3})$ time complexity operation can be computationally demanding especially 
 when the number of points is large. 
Therefore, as motivated in \cite{karni2000spectral}, we resort to mesh partitioning, where we segment 
 a mesh into smaller constituents and compress each segment independently. 
We use the METIS \cite{karypis1998fast} package for fast graph partitioning and segmentation. 
Therefore, the time complexity of the proposed framework can be seen as \emph{linear} in the number of partitions of fixed size.

The Neumann boundary conditions induced by the graph Laplacian may cause a discrepancy between the reconstructed segments along shared boundaries. Mesh partitioning may result in a so-called \emph{edge-effects} \cite{karni2000spectral}, a degradation of the reconstruction along the sub-mesh boundaries.  However, compressing smaller meshes separately has the advantage of better capturing local properties of the mesh even with low number of coefficients. Also, from our experiments, this degradation is not visually sensitive even for low compression rate and especially not with the suggested method.
We partition the mesh into sub-meshes of few hundred vertices. This ensures both faster computation of the coefficients and fewer difficult areas for reconstruction the Hamiltonian operator can focus on.

\subsection{Results}
\begin{figure*}[tbp]
  \centering
  \mbox{}
  \includegraphics[width=\linewidth,trim={0 80 0 0},clip]{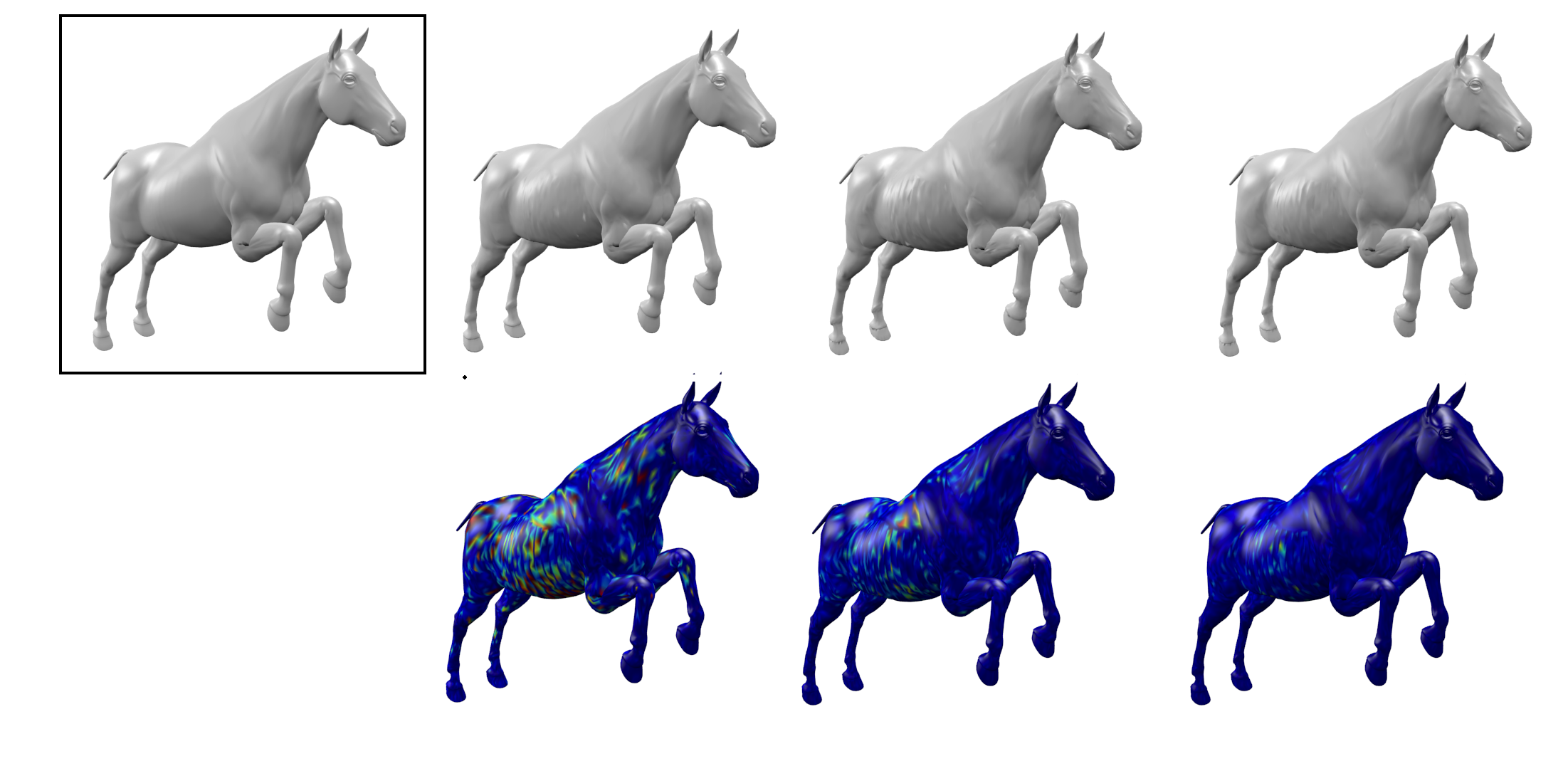}
  \hfill \mbox{}
  \caption{\label{fig:horseRecon}
Horse model (boxed mesh) composed of 19248 vertices. (Top) Shape reconstruction for a compression ratio of $4:10$ using dictionaries composed of the MHB, MHB+SGW and MHB+Hamiltonian (from left to right).
(Bottom) Reconstruction error using the proposed spectral bases.
 }
\end{figure*} 

Figures \ref{fig:fandiskRecon} and \ref{fig:horseRecon} show a visual comparison of the shape approximation results. We propose two visualizations. The first row presents a regular reconstruction with the three different methods for a fixed compression ratio. The second row shows a profile of the error plot over the original mesh to highlight regions of erroneous representation. Clearly the MHB has a fairly large error profile over the surface. The results with the spectral graph wavelets show an improvement as compared to the manifold harmonics. However, the sparse approximation with the Hamiltonian basis shows a distinct positive difference as compared to the other two methods both in the reconstruction and error profiles. 

\begin{figure}[htbp]
\begin{overpic}[width=1\linewidth,trim={350 30 430 35},clip]{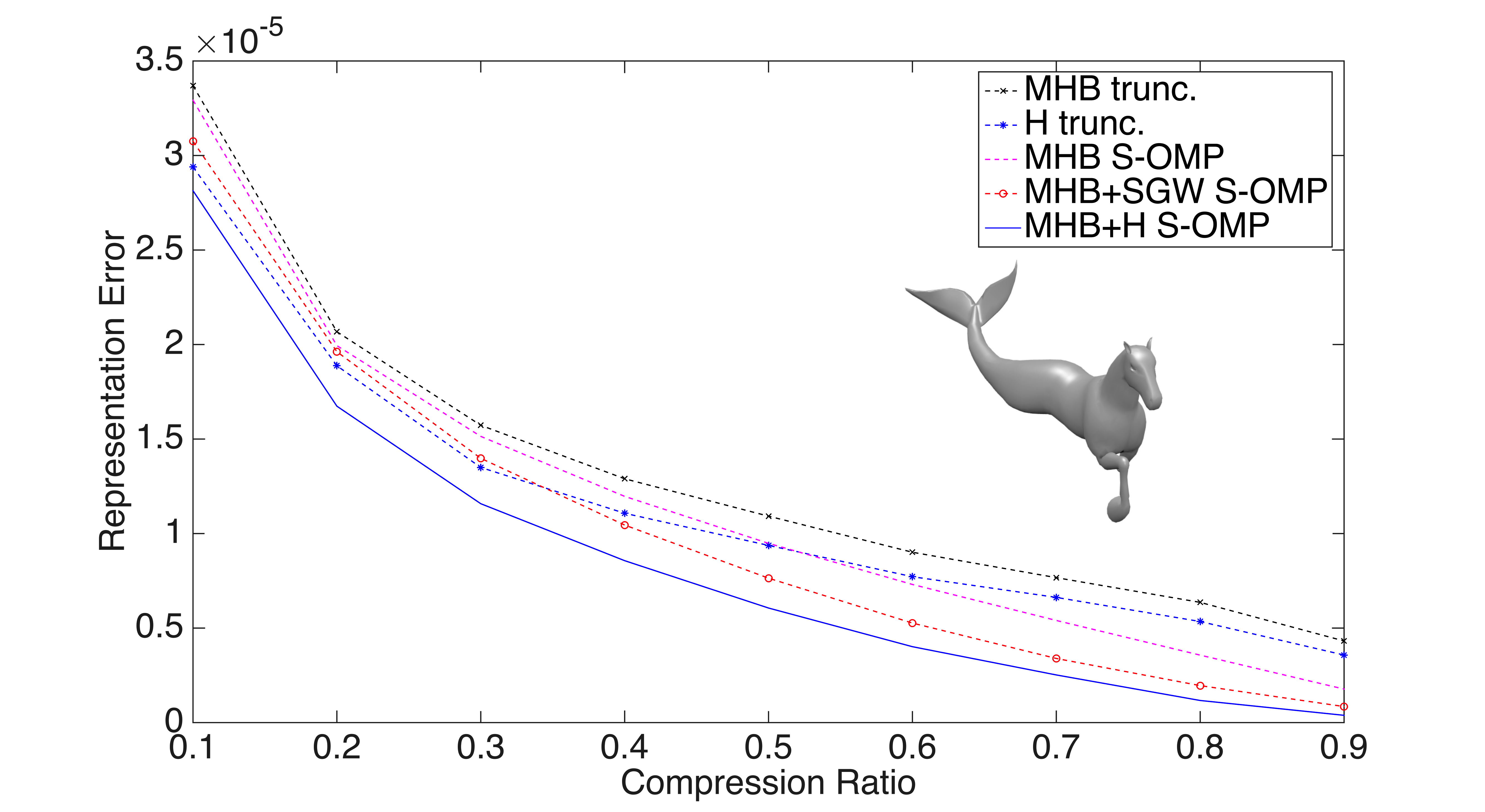}
\centering
 \end{overpic}
 \caption{\label{fig:seaHorse} 
 Reconstruction error as a function of the compression ratio for the seahorse model (2194 vertices). }
 \end{figure}

\begin{figure}[htbp]
\begin{overpic}[width=1\linewidth,trim={350 30 430 35},clip]{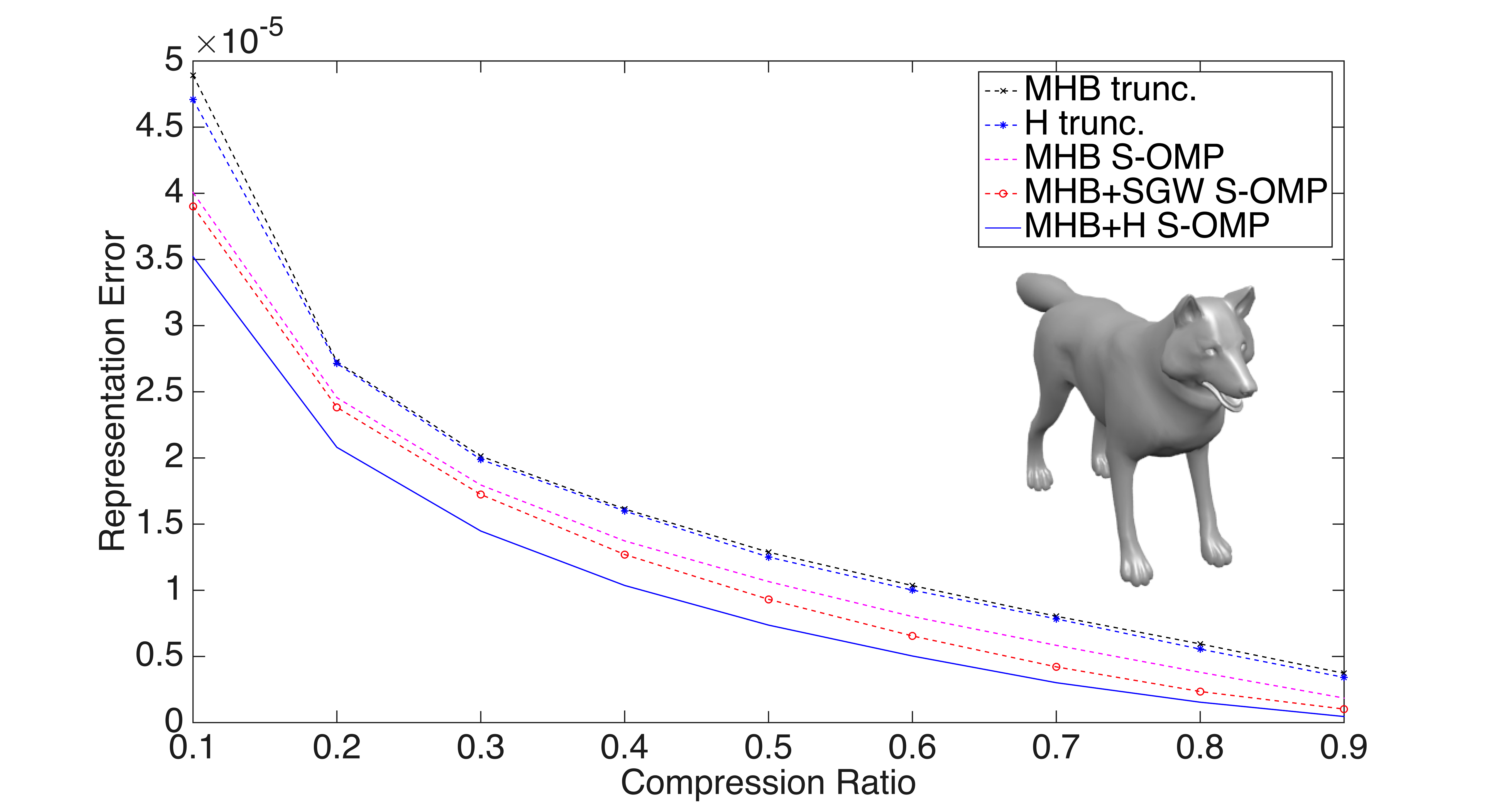}
\centering
 \end{overpic}
 \caption{\label{fig:wolf} 
 Reconstruction error as a function of the compression ratio for the wolf model (4344 vertices).}
 \end{figure}

\begin{figure}[htbp]
\begin{overpic}[width=1\linewidth,trim={350 30 430 35},clip]{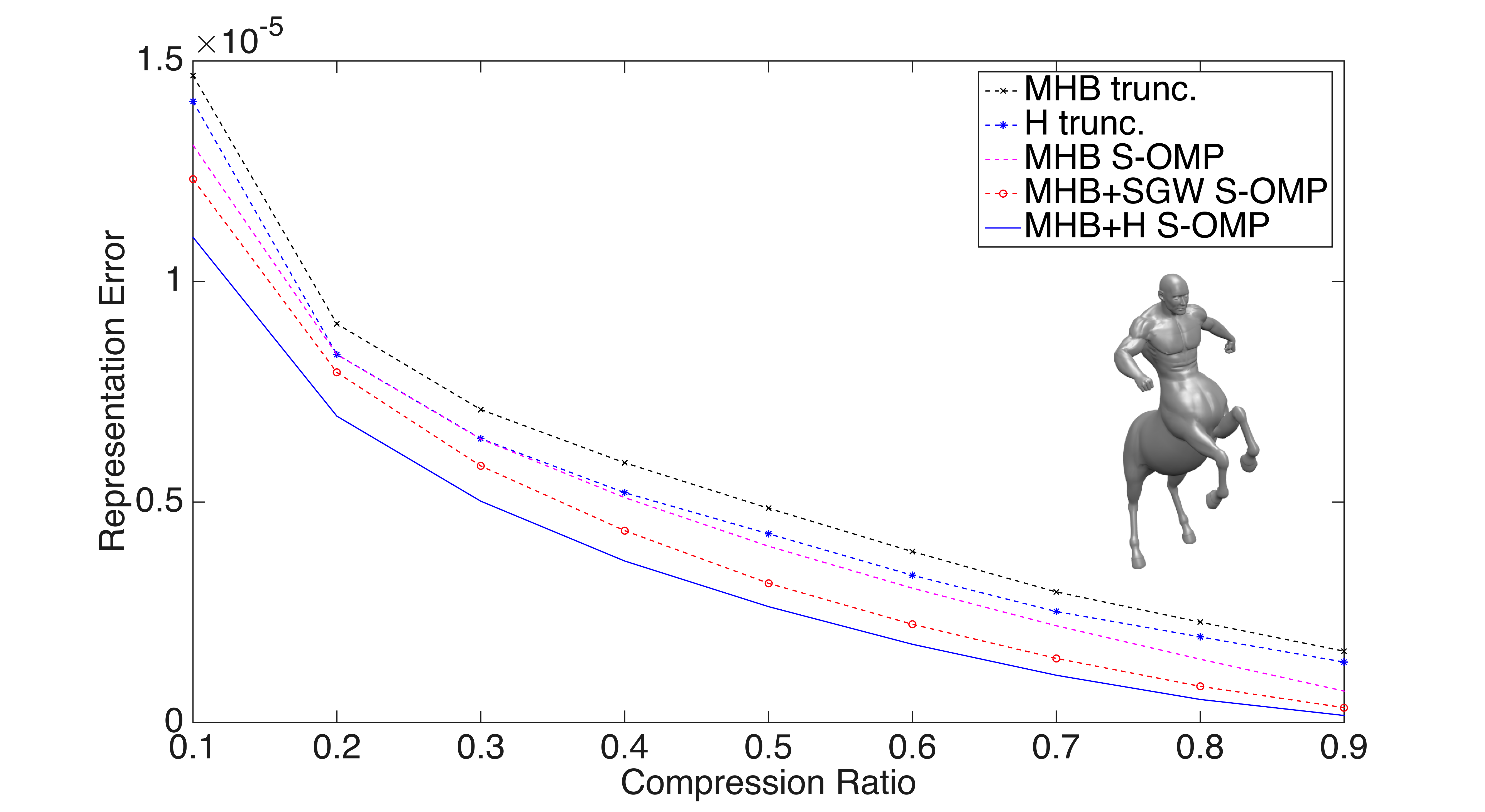}
\centering
 \end{overpic}
 \caption{\label{fig:centaur} 
 Reconstruction error as a function of the compression ratio for the centaur model (15768 vertices).}
 \end{figure}

\begin{figure*}[ht]
  \centering
  \mbox{}
  \includegraphics[width=\linewidth,trim={0 0 0 0},clip]{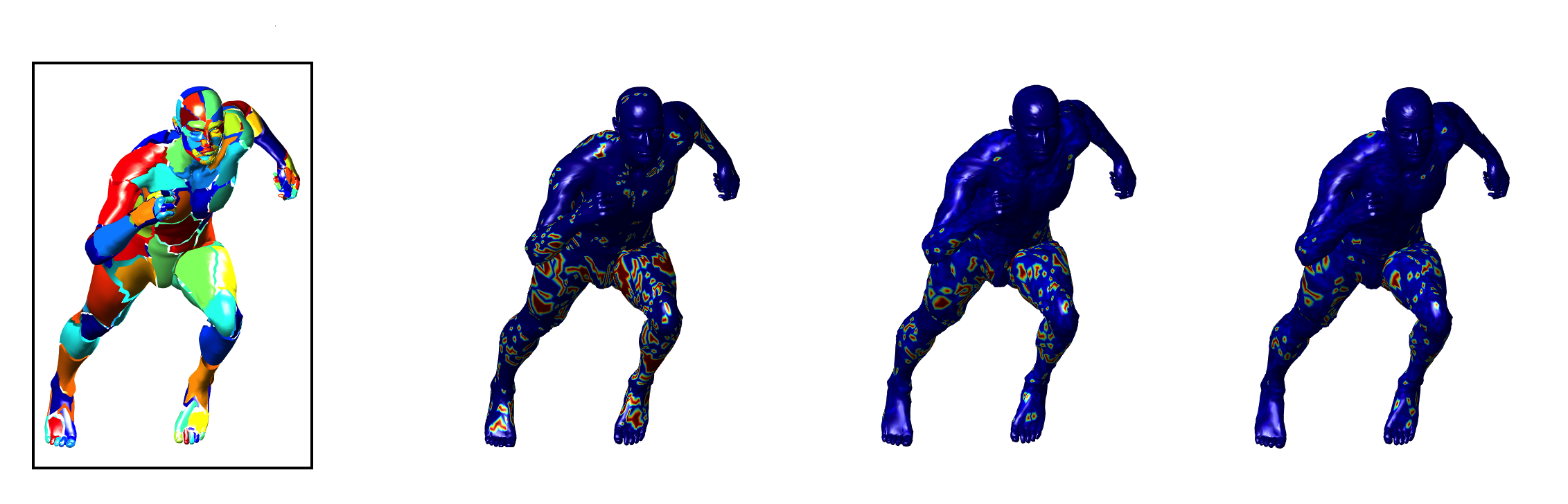}
  \hfill \mbox{}
  \caption{\label{fig:segments}
Michael model after mesh partition (boxed mesh)  composed of 52565 vertices followed by the shape reconstruction error map for a compression ratio of $1:10$ using MHB truncation and dictionaries composed of the MHB+SGW and MHB+Hamiltonian (from left to right).
The error scale of the the MHB truncation is ten times bigger than the others.
 }
\end{figure*}

Figures \ref{fig:seaHorse}, \ref{fig:wolf} and \ref{fig:centaur} show compression results of our
 method compared with other spectral techniques on different shapes.
We present results according to four prominent spectral methods.  
First with the plain truncation scheme of the Manifold Harmonic Basis (MHB) as suggested in \cite{karni2000spectral}. 
Second, the truncation scheme of the Hamiltonian with optimal potential as suggested in \cite{ChoukrounSK16}. 
Third, MHB with sparse coding i.e. using the S-OMP algorithm (MHB-SOMP) on the manifold harmonics dictionary.
Fourth, we present Spectral graph wavelets with sparse coding (MHB+SGW-SOMP) using dictionaries defined with the
 harmonics and the graph wavelets as presented in \cite{zhong2014sparse}.We used 32 bits as for single bit-precision in the computation of compression ratio.
 An illustration of the mesh partition result and reconstruction performance between the methods is present in Figure \ref{fig:segments}.

We can see that our algorithm shows a considerable improvement in compression performance by generating smaller errors over the entire span of compression requirements for different shapes, even under non-uniform triangulations. 
We highlight the fact we leverage benefits in optimizing on both fronts: a superior basis and the improved sparse coding mechanism for representation. Using only the Hamiltonian basis in a simple truncation scheme as suggested in \cite{ChoukrounSK16} shows improvement over standard spectral techniques making it comparable to the pursuit based solutions, but the major gain comes only after its use in conjunction with sparse representation.

Finally, in order to gain a global assessment of the performance of our method in comparison to generic non-spectral mesh compression algorithms (motivated from \cite{maglo20153d}), 
we compiled an average bits-per-vertex (bpv) analysis in table \ref{table:1}, where the proposed method bpv is obtained on the same shapes tested in \cite{karni2000spectral} and with the same reconstruction error threshold.
However, since other prominent non-spectral methods like \cite{khodakovsky2000progressive,guskov2000normal} involve some form of remeshing, the comparison may not be entirely appropriate, but provides a high level comparison between the compression types.
The general comparison shows that spectral algorithms show lower compression rates as compared to methods like \cite{khodakovsky2000progressive} and \cite{guskov2000normal}. However, our approach of using the spectral route along with sparse representations shows considerable gains without any any assumption on the mesh connectivity.

 \begin{table}[h!]
\begin{tabular}{ | m{3cm}|| >{\centering}m{2cm}| m{2cm}|}
 \hline
{\bf Algorithm} & {\bf Total Compression rates (bpv)} & {\bf Compression Type}\\
 \hline
 Progressive Meshes \cite{hoppe1996progressive}   & 37    &   Vertex split\\
  \hline
 Wavemesh \cite{valette2004wavelet}&   19     & Irregular Wavelets\\
  \hline
 Spectral Compression \cite{karni2000spectral}&   19     & Spectral\\
  \hline
   Incremental parametric refinement \cite{valette2009progressive}&   15     & Parametric refinement\\
  \hline
 \bf{Spectral Hamiltonian} (proposed method) &   \bf{12 }    & Spectral\\
 \hline
 Wavelet Compression \cite{khodakovsky2000progressive} &   8 & {\emph{Remeshing based}}\\
 \hline
 Normal Meshes \cite{guskov2000normal} &   6 & {\emph{Mesh-Sequence compression}}\\
 \hline
\end{tabular}
\caption{Comparison of the suggested method with the main progressive mesh compression algorithms.}
\label{table:1}
\end{table}
 
\section{Conclusion and Future Work}
We introduced the Hamiltonian operator for spectral mesh compression. 
This operator uses the concept of a potential function whose action leads to the modulation
 of the spectral geometry of that domain. 
We use this operation to modify the standard manifold harmonics of the 3D surface by emphasizing 
 the role of regions with errors that emerge as a result of compression with classical harmonics.
We maximize compression performance by constructing meaningful dictionaries from this basis
 in a sparse approximation framework and show a distinct improvement from previous spectral techniques.

In future work, we intend to explore the design and construction of newer meaningful potential
 functions for 3D surfaces.
The use of the \SCH operator for generic graph structures or discrete manifolds is an unchartered
 territory that can lead to interesting spectral embeddings.
\nocite{sgp.20171205,DBLP:journals/corr/ChoukrounSK16,sgp.20171203,10.1111:cgf.13252}
\bibliographystyle{spmpsci}      
\bibliography{references2.bib}   

\end{document}